\pdfoutput=1

\documentclass[11pt]{article}

\usepackage[final]{acl}

\usepackage{times}
\usepackage{latexsym}
\usepackage{alltt}
\usepackage{fvextra}
\DefineVerbatimEnvironment{Verbatim}{Verbatim}{breaklines=true}

\usepackage[T1]{fontenc}

\usepackage[utf8]{inputenc}

\usepackage{microtype}

\usepackage{inconsolata}

\usepackage{graphicx}
\usepackage{booktabs}
\usepackage{amsmath}
\usepackage{subcaption}
\usepackage{multirow}

\usepackage{multirow}
\usepackage{siunitx}
\title{Open Domain Question Answering with Conflicting Contexts}

\author{
 \textbf{Siyi Liu \thanks{Work done during internship at AWS AI Labs} \textsuperscript{1,2}},
 \textbf{Qiang Ning\textsuperscript{1}},
 \textbf{Kishaloy Halder\textsuperscript{1}},
 \textbf{Wei Xiao\textsuperscript{1}},
\\
 \textbf{Zheng Qi\textsuperscript{1}},
 \textbf{Phu Mon Htut\textsuperscript{1}},
 \textbf{Yi Zhang\textsuperscript{1}},
 \textbf{Neha Anna John\textsuperscript{1}},
\\
 \textbf{Bonan Min\textsuperscript{1}},
 \textbf{Yassine Benajiba\textsuperscript{1}},
 \textbf{Dan Roth\textsuperscript{1,2}}
\\
\\
 \textsuperscript{1}AWS AI Labs,
 \textsuperscript{2}University of Pennsylvania,
\\
 \texttt{
 siyiliu@seas.upenn.edu
 }
}

\begin{document}
\maketitle
\begin{abstract}
Open domain question answering systems frequently rely on information retrieved from large collections of text (such as the Web) to answer questions. However, such collections of text often contain conflicting information, and indiscriminately depending on this information may result in untruthful and inaccurate answers. To understand the gravity of this problem, we collect a human-annotated dataset, \textit{Question Answering with Conflicting Contexts (QACC)}, and find that as many as $25\%$ of \textit{unambiguous}, open domain questions can lead to conflicting contexts when retrieved using Google Search. We evaluate and benchmark three powerful Large Language Models (LLMs) with our dataset \textit{QACC} and demonstrate their limitations in effectively addressing questions with conflicting information. To explore how humans reason through conflicting contexts, we request our annotators to provide explanations for their selections of correct answers. We demonstrate that by finetuning LLMs to explain their answers, we can introduce richer information into their training that guide them through the process of reasoning with conflicting contexts. We publicly release our dataset and code to promote research along this line\footnote{\url{https://github.com/amazon-science/qa-with-conflicting-context}}.

\end{abstract}

\section{Introduction}

Large language models (LLMs) have shown impressive capabilities on question answering tasks. In an open domain setting, a typical approach involves (1) retrieving relevant documents as contexts from the web or knowledge bases, and (2) using LLMs to generate the answer with the guide of the context. However, retrieved contexts from the web could often present \textbf{conflicting} information: \textit{e.g.,} $22.62\%$ pregnant women reported to find conflicting medical information from different websites in a survey \cite{info:doi/10.2196/jmir.2939}, such conflicts can lead to undesirable consequences when a language model relies indiscriminately on them to answer questions.

Previous work has explored different aspects of conflicts in the field of Natural Language Processing (NLP), including having different perspectives \cite{chen2019seeing, liu2021multioped}, fake news and misinformation \cite{chen2022design, pan2023attacking}, conflicts due to ambiguous or inadequate questions \cite{min2020ambigqa, SituatedQA}, knowledge that changes over time \cite{NEURIPS2023_9941624e}, and conflicts between knowledge encoded in the parameters and provided in the contexts \cite{longpre-etal-2021-entity, chen-etal-2022-rich, xie2024adaptive}.

\begin{figure}
  \includegraphics[width=7.9cm]{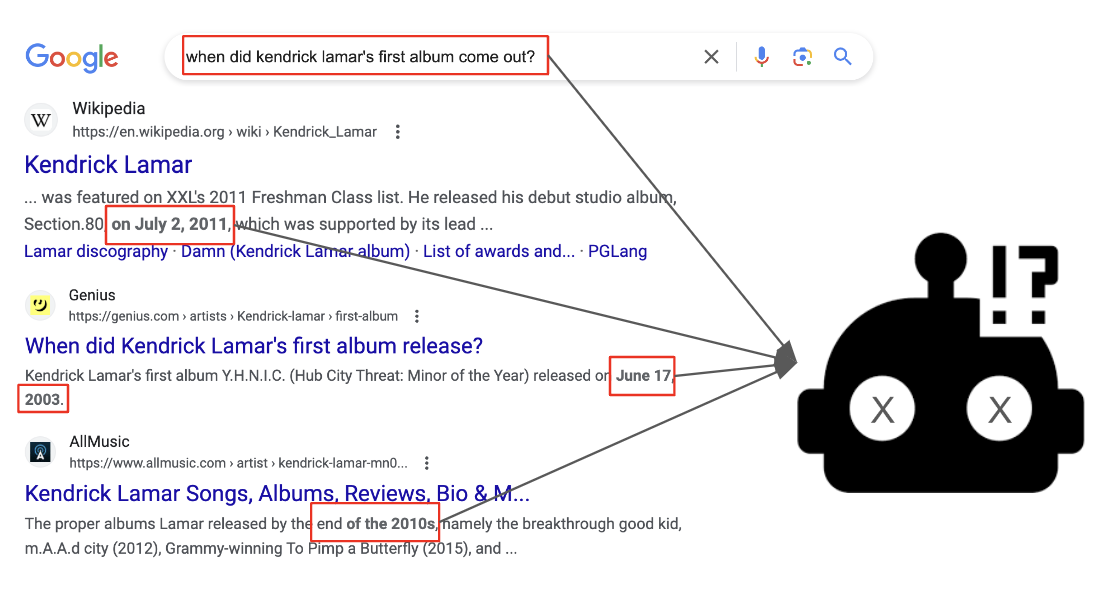}
  \caption{Google search results when querying the question \textit{"When did Kendrick Lamars first album come out?"}. We can see that here different answers (July 2, 2011 / June 17, 2003 / end of the 2010s) are suggested by Google and it is difficult for a language model to decide which to believe in.}
  \label{fig:google}
\end{figure}

In this work, we target the \textbf{conflicts among contexts} when retrieving from the web with an \textit{unambiguous} query and study their impact on the downstream question answering task.
Figure \ref{fig:google} shows the results of querying \textit{"when did kendrick lamars first album come out?"} on Google\footnote{Results were queried in June 2024.}. We observe that in the top-$10$ returned results, there is evidence suggesting different answers to the question, and such inconsistencies may confuse the language models when they refer to the contexts to answer the question. 
Earlier work studied this issue of conflicting contexts through perturbations with entity-substitution \cite{chen-etal-2022-rich, hong2024gullible}, machine-generation \cite{pan2023attacking, wan2024evidence, hong2024gullible}, rule-based templates \cite{kazemi2023boardgameqa}, or on controversial, multi-perspective questions \cite{liu2021multioped, wan2024evidence}. However, none of them examine the scenario where realistic, unambiguous open domain questions can also lead to conflicting contexts on the web and its effect in downstream question answering.

To quantify how often conflicting contexts occur on the web,
we construct our dataset named QACC (\textbf{Q}uestion \textbf{A}nswering with \textbf{C}onflicting \textbf{C}ontexts). We consider \textit{unambiguous} open domain questions from AmbigQA \cite{min2020ambigqa}
and use Google Search API\footnote{\url{https://developers.google.com/custom-search/v1/overview}} to retrieve up to $10$ results for each question. We then use Amazon Mechanical Turk and ask human annotators to determine whether there exists different answers in the contexts. We find that about $25\%$ of the \textit{unambiguous} open domain questions will yield conflicting evidence from Google. We evaluate three popular LLMs (GPT-4o, Claude-3, and Phi-3) on our dataset with different prompting and finetuning strategies and establish that conflicting contexts can lead to substantial performance degradation in them. To understand how humans reason through conflicting contexts, we ask our annotators to select from a pre-defined set of reasons when deciding on the answer. Our findings indicate that humans often adhere to majority vote (i.e. selecting the most popular answer) when seeing conflicting contexts. In addition, we also request our annotators to provide a single sentence, natural language explanation for their answers. We find that by finetuning LLMs to explain their answers, we can introduce richer information into their training that guide them through the process of reasoning with conflicting contexts and improve their performance in both QACC and a perturbed NQ-Open dataset \cite{lee-etal-2019-latent}.

To summarize, our contributions in this work are the following:

\begin{itemize}
    \item We construct a human-annotated dataset QACC and find that about $25\%$ of unambiguous, open domain questions can lead to conflicting contexts when queried with Google Search.
    \item We benchmark open domain question answering with conflicting contexts with our dataset QACC and demonstrate the limitations of current LLMs under this scenario.
    \item We show that when finetuning with human explanations, LLMs can improve their abilities to answer questions correctly with conflicting contexts.
\end{itemize}

\begin{figure*}
  \includegraphics[width=\textwidth]{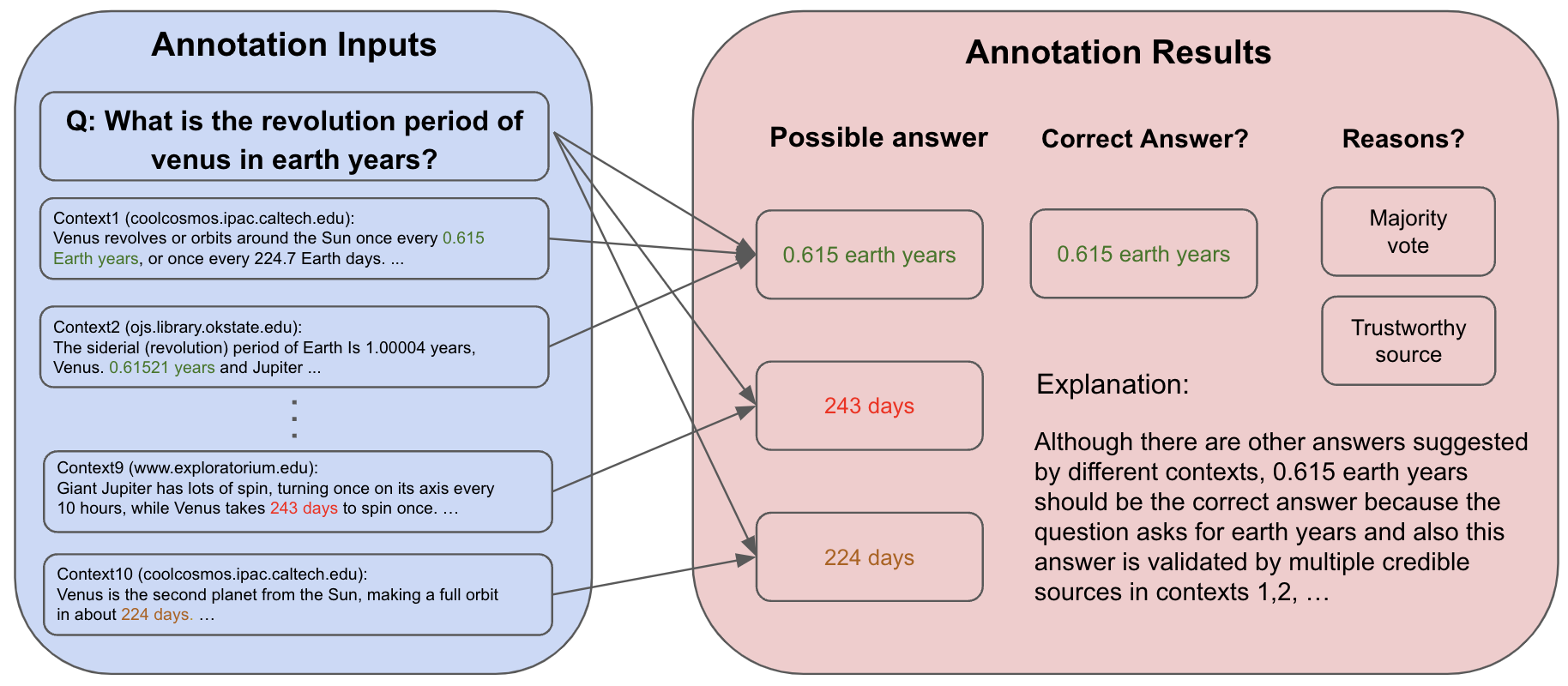}
  \caption{Data Collection Pipeline. The left side shows an example input given to the annotators, and the right side shows an example annotated result. During the annotation process, we ask the annotators to identify different possible answers given each context, and decide there is a conflict if there is more than one possible answer. In addition, we ask the annotators to select from a pre-defined list of reasons, and provide a natural language explanation of their decision. In this example, the annotators believe there are three possible answers, and they think \textit{0.615 earth years} is the correct answer because it's validated by most trustworthy sources. }
  \label{fig:data_collection}
\end{figure*}

\section{Related Work}
\subsection{Retrieval Augmented Question Answering}

Open-domain question answering (ODQA) aims to answer factoid questions with a large collection of documents \cite{voorhees-tice-2000-trec}. With the new advances in large language models (LLMs), a typical approach to OPQA involves a two-stage framework: (1) first retrieve a small subset of passages where some of them contain the answer to the question, and then (2) use a LLM to answer the question using the retrieved passages as contexts \cite{chen2017reading, karpukhin2020dense, guu2020realm, khandelwal2020generalization, izacard2021leveraging, borgeaud2022improving, zhong2022training}. Retrievers augment question answering by retrieving up to 100 passages and set new state-of-the-art for ODQA \cite{izacard2021leveraging}; however, we believe that with such large amount of passages retrieved as context, it's frequent for them to contain conflicting information, and such conflicts will confuse the downstream language models in question answering. In this work, we validate this hypothesis and show that for a retriever like Google Search, $25\%$ of the time it will return conflicting contexts in its top ten results when queried with a realistic, unambiguous question. We further demonstrate the limitations of current LLMs under this scenario of conflicting contexts through our experiments.


\subsection{Knowledge Conflicts}
\paragraph{Parametric v.s. Contextual}
One line of work studies knowledge conflicts in the setting of parametric v.s. contextual knowledge. Parametric knowledge refers to the knowledge a model learns during pre-training, and contextual knowledge refers to the contextual information a model sees at inference time. \citet{longpre-etal-2021-entity} proposes a entity-substitution framework that identifies QA instances with named entity answers and then substitutes mentions of the entity in the gold document with an alternate entity to create entity-based knowledge conflicts. \citet{chen-etal-2022-rich} expands the study to consider multiple evidence passages and shows that when some passages are perturbed not to support an answer, language models largely ignore semantic perturbations and outputs potential answer entity in the retrieved passages. \citet{xie2024adaptive} proposes another new framework to elicit the parametric memory of LLMs in order to construct the corresponding counter-memory and shows that with both supportive and contradictory evidence to their parametric memory, LLMs show a strong confirmation bias and tend to cling to their parametric memory. \cite{liu2024untangle} proposes a machine-generated dataset of knowledge conflicts and studies different strategies to enable LLMs to resolve conflicting knowledge.

\paragraph{Contextual v.s. Contextual}
Another line of work focuses on the scenario when a language model is given conflicting contexts as in our setting. Some previous work create conflicting contexts with perturbations, including entity-substitution \cite{chen-etal-2022-rich, hong2024gullible} and machine-generation \cite{pan2023attacking, wan2024evidence, hong2024gullible}, and some other work define conflicts over rule-based templates \cite{kazemi2023boardgameqa}. However, most of these previous work are built on synthetic data, whereas our dataset are real-world search results of unambiguous questions. \citet{wan2024evidence} also uses Google Search to extract conflicting contexts, but they focus specifically on controversial and contentious questions and analyze the linguistic features in the text that affect language models' predictions, whereas we show that realistic, unambiguous questions can also lead to conflicting contexts from Google and finetuning LLMs on our human written explanations can teach them to reason through the conflicts.

\section{Question Answering with Conflicting Contexts}
In this section, we discuss our exploration of the problem: question answering with conflicting contexts. We first suggest our definition of what constitutes as conflicting context, then introduce how we collect a our dataset QACC for the analysis, and lastly share our findings and analysis of QACC.

\subsection{Problem Definition}
Given a question $q$, a list of retrieved contexts $C=\{c_1, c_2, ..., c_i\}$, and a question answering system $\phi$, we can get a list of individual answers $A=\{a_1, a_2, ..., a_i \}$, where $a_i= \phi(q, c_i)$. We state that the question $q$ has conflicting evidence if and only if $\exists(a_i, a_j \in A ) (a_i \neq a_j)$. In other words, at each step a question answering system (a human or a language model) is given the question and only one of the context in order to answer the question. Iterate this through all of the contexts, and we state that there are conflicting contexts if and only if there are different answers generated when given different contexts. 

Our research problem is constrained to factual, unambiguous questions. Specifically, each question in our dataset is expected to have a single, definitive factual answer. Subjective questions, unanswerable questions, or questions with multiple answers are beyond the scope of this work.
In addition, note that our definition of \textit{conflict} here differs from the definition of \textit{contradiction} in traditional Natural Language Inference (NLI) tasks. Here we define \textit{conflict} in a way that is less restricting than \textit{contradictory} texts in NLI to further exploit its applicability in our target domain, open domain QA with Retrieval-Augmented Generation (RAG). For instance, in a RAG scenario, retrievers can often retrieve contexts that contain seemingly correct answers (e.g. \textit{neutral} texts in the case of NLI), and such different/conflicting answers may also confuse the downstream LLMs. We believe that our definition of \textit{conflict} and the resulted dataset can therefore better support us towards our goal. 

\subsection{QACC Dataset}

\textit{Ambiguous} questions can frequently lead to multiple different answers \cite{min2020ambigqa}. However, we believe that even when questions are \textit{unambiguous}, it is still common to see conflicting evidence on the web. To this end, we consider AmbigQA \cite{min2020ambigqa}, a dataset with questions labeled as either \textit{ambiguous} or \textit{unambiguous}, and take only questions that are labeled as \textit{unambiguous} as the the questions in our dataset.  We then use Google Search API to retrieve top-10 search results as the contexts for each question, and use Amazon Mechanical Turk to collect annotations for each question and its associated contexts. The statistics of our dataset QACC is shown in Table \ref{tab:stats}, and QACC is in English language.

\subsection{Human Annotation}
We employ a rigorous human annotation pipeline with a qualification exam before the main annotation task, a strategy commonly used to ensure the collection of high-quality datasets \cite{han-etal-2021-ester, dasigi2021dataset}. Only annotators that have passed our qualification exams can participate in the main annotation task. We use Amazon Mechanical Turk (MTurk) to collect the annotations and CROWDAQ to design the annotation interface  \cite{ning-etal-2020-easy}. All of our annotators were MTurk workers who self-reported as being located in the U.S., and we specifically hired workers with the “Masters” qualification\footnote{Amazon Mechanical Turk award workers master's qualifications only if they have demonstrated superior performance over a period of time across thousands of annotations.} on MTurk. Other demographic details about the annotators were unspecified and unavailable.
To ensure fair compensation, we adhered to the U.S. Federal Minimum Wage of 7.25 US Dollar per hour. Initially, we launched a pilot batch of annotations, compensating annotators 0.5 US Dollar per task. From this pilot, we observed that the average time to complete an annotation task was 2 minutes and 29 seconds. Based on this estimate and the federal minimum wage, we adjusted the compensation to 0.35 US Dollar per annotation task for the rest of the annotations. The average time spent on these tasks was later measured at 2 minutes and 49 seconds, supporting our conclusion that the compensation rate aligns with fair pay standards. Each question is annotated by one annotator, and examples of our annotation instruction, qualification and annotation interfaces are shown in Appendix \ref{appendix:annotation}.

\paragraph{Qualification Exam}
Since the annotation task requires critical thinking and an attention to detail, we design interactive tutorials and request the annotators to review them before the qualification exams. We first show them instructions and our definition of conflicting contexts for a question, and then ask them to complete a set of tutorial questions where we display the expected answers and reasons once they answer them.
After they understand the goals and formats of the annotations, we request them to complete a set of 12 random, multiple-choice qualification questions. Only workers with more than $90\%$ accuracy on the exam can pass and get the qualification to participate in our main annotation task. We allow only the workers that have a master's qualification to take the exam, and 12 among 41 of them ($29\%$) have passed our exam and participate in our main annotation.

\paragraph{Main Annotation}
Following our definition of the problem, we ask the annotators to identify the conflict in the contexts by finding different possible answers. In each Human Intelligence Task (HIT), we show the annotator an open domain question, a list of contexts retrieved by Google Search, as well as the website domains these contexts are from. We then ask the annotators the following questions: 1. Is there more than one possible answer when looking at the question and each context individually (conflict identification)? 2. Which of the contexts support which of the different answers (answer attribution)? 3. Which answer do you think is the correct answer (question answering)? 4. why do you think the answer you choose is correct (QA with explanation)?. The fourth question here includes both a multiple-choice question that asks them to select a reason from a pre-defined set and a free-form question that asks them to explain their reasoning in a single sentence.
These procedures result in a rich annotation of QACC that can also support other QA-related tasks not covered in the scope of this work, like answer attribution. An example of the dataset and the data collection process is shown in Figure \ref{fig:data_collection} in Appendix.

\begin{table}[]
\begin{tabular} {ccc}
\hline
Split  & \# of QAs & \% QAs with Conflicts \\
\hline
Train & 394 & 29\%  \\

Dev & 303 & 19\%  \\

Test & 813 & 25\%  \\

\hline
\end{tabular}
\centering
\caption{The statistics of QACC.}
\label{tab:stats}
\end{table}

\subsection{Expert Verification}

To evaluate the quality of the annotations, one of the authors randomly selects 50 examples from QACC for expert verification. A prior internal pilot study with expert annotations demonstrated high agreement among five team members, supporting the use of a single author for this verification process. We compare our annotations with those of the annotators across two tasks: 1. determining whether there is a conflict, and 2. assessing whether our suggested correct answer aligns with that of the annotators. We exclude the reasons our annotators selected from the predefined list and the annotators' natural language explanations from inter-annotator agreement assessment, as these aspects involve subjective human judgments, which can naturally vary among individuals. Comparing our annotations with those provided by MTurk workers, we observe a Cohen’s kappa agreement of 0.615 for the first task, and find that 88\% of our suggested correct answers align with those of the annotators, indicating a high level of annotation quality in our dataset.

\begin{figure}
  \includegraphics[width=8cm]{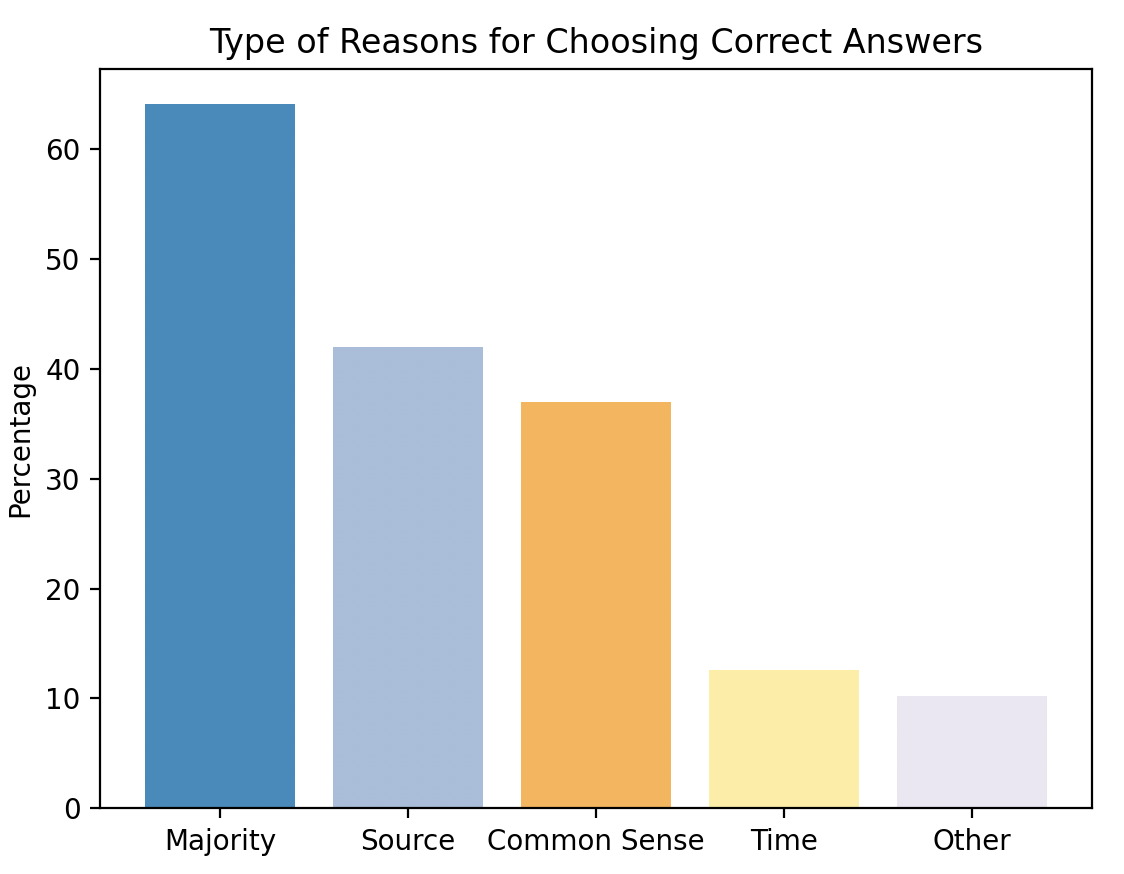}
  \caption{Reasons of annotators selecting one correct answer over the others when there are conflicts. "Majority" means the answer is supported by the most contexts. "Source" means the annotator trust the contexts more because they come from trustworthy sources. "Common Sense" means the answer matches their own memory and common sense. "Time" means they think one answer is correct since it's the most up-to-date.}
  \label{fig:data_analysis}
\end{figure}

\subsection{Statistics and Analysis}

Table \ref{tab:stats} shows the statistics of our dataset. We can see that about $25\%$ of all the unambiguous, open domain questions in our dataset have conflicting contexts when retrieved using Google Search. Among the questions identified as having conflicting contexts, the average number of distinct answers is $2.47$. Specifically, $25\%$ of all questions exhibit conflicts (i.e., they have at least two different answers), $10\%$ have at least three distinct answers, and $3\%$ have at least four. Furthermore, within the subset of questions with conflicting contexts, $29\%$ have one answer supported by at least half of the contexts.

To better understand humans' reasoning process when presented with conflicting evidence, we ask the annotators to choose from a pre-defined list of reasons that can best categorize why they think one of the answers is correct. We allow them to choose more than one option since different factors can simultaneously affect one's decision in choosing the correct answer. Figure \ref{fig:data_analysis} shows their reasons when the question is labeled by them as having conflicting contexts. We find that humans favor answers that are the most popular in the contexts the most, and also refer to the sources of the context (trustworthy or not) and their own intuitions and common sense about the question when deciding on the answer. On the other hand, fewer annotators select correct answers based on the time the information was published.

\begin{figure}
  \includegraphics[width=8cm]{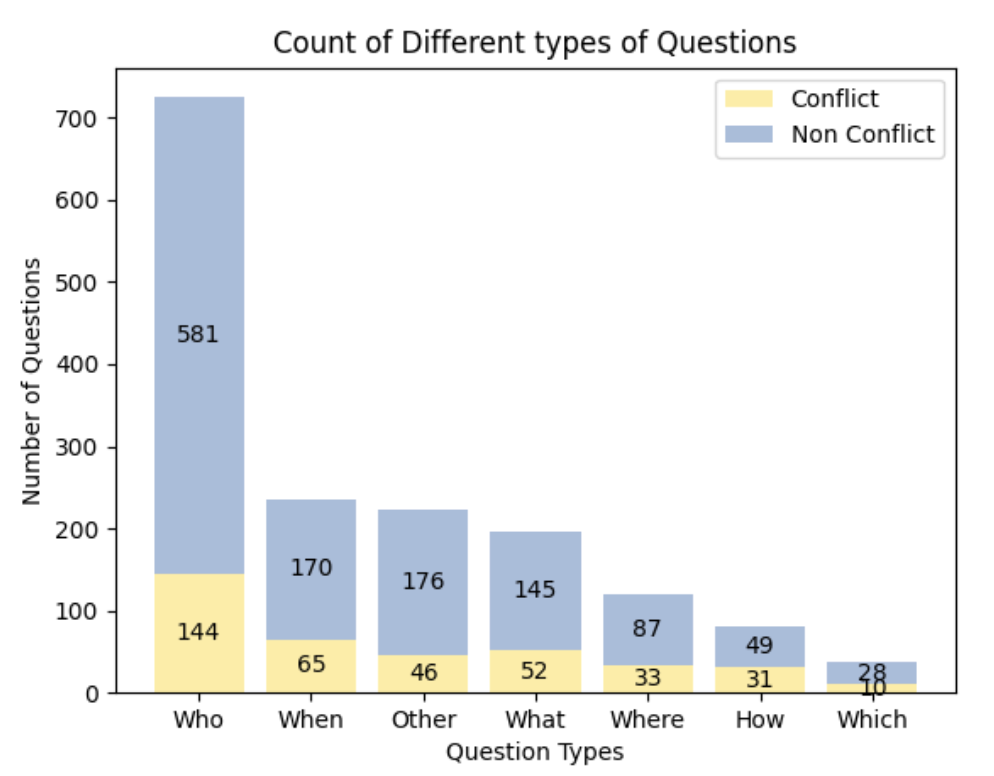}
  \caption{Different types of questions in our dataset that have conflicts.}
  \label{fig:type_analysis}
\end{figure}

We also conduct data analysis to study the different types of questions in our dataset that lead to conflicting contexts.
In Figure \ref{fig:type_analysis}, we see that most of the questions in our dataset are \textit{Who} questions. This type of question has about $20\%$ of times that lead to conflicting contexts. \textit{How} questions, on the other hand, can lead to conflicting contexts almost $40\%$ of the time.
This aligns with our hypothesis since questions that start with \textit{How} are typically open-ended questions with a more complex answer and can involve different perspectives. 

Another characteristic about QACC that we observe is that the most prominent type of conflicts between different answers is the mismatch of entities, i.e., different names, time, and places. This is largely due to the fact that we source our questions from AmbigQA and Natural Questions, where the expected answers are short phrases that are mostly entities.


\section{Experiments}
In this section, we benchmark the problem of question answering with conflicting contexts with three popular LLMs. We demonstrate that teaching language models to explain its answer can guide their inference process and improve their performance in QACC, and such improvement can generalize to another perturbed NQ-Open dataset.

\begin{table}[t]
\begin{tabular} {c|cc}
\hline
Model  & \textsc{EM} & \textsc{F1}  \\

\hline
Few-shot Exp \& Ans & 52.64 & 66.84  \\
Finetuned Exp \& Ans & \textbf{54.74} & 67.19  \\

Few-shot Ans \& Exp & 51.41 & 65.78  \\
Finetuned Ans \& Exp & 53.75 & \textbf{67.24}  \\

\hline
\end{tabular}
\centering
\caption{Experiments on the order of explanation and answer on Phi-3-Medium evaluated on QACC. We observe only slight differences in terms of the performance.}
\label{tab:ablation}
\end{table}

\begin{table*}[t]
\begin{tabular} {c|c|ccc|ccc}
\hline

Model & Prompt & \textsc{EM-C} & \textsc{EM-NC} & \textsc{EM-T} & \textsc{F1-C} & \textsc{F1-NC} & \textsc{F1-T}\\
\hline
\hline


\multirow{4}{*}{GPT-4o-few-shot} &  Context & \underline{42.51} & 54.13 & \underline{51.17} & \underline{59.29}& \underline{\textbf{70.06}} & \underline{\textbf{67.32}} \\
  
&  + Majority \& Ans & 40.58 & 53.96& 50.55 & 57.35& 69.03 & 66.05 \\

  &  + Discern \& Ans & 39.13 & 52.15& 48.83 & 56.33& 67.81 & 64.88 \\

 &  + Exp \& Ans & 38.65 & \underline{54.29}& 50.31 & 57.43& 69.19 & 66.2 \\

\hline

\multirow{4}{*}{Claude-3-few-shot} & Context & 35.75 & 53.14 & 48.71 & 53.76 & 67.9 & 64.3 \\
 &  + Majority Vote & 37.68 & 52.64 & 48.83 & 54.75 & 67.85 & 64.51 \\
 &  + Dis \& Ans & 31.4 & 43.89 & 40.71 & 48.52 & 58.48 & 55.95 \\
 &  + Exp \& Ans & \underline{40.1} & \underline{54.29} & \underline{50.68} & \underline{57.41} & \underline{69.28} & \underline{66.26} \\
\hline
 \multirow{4}{*}{Phi-3-few-shot}& Context & 42.51 & 51.98 & 49.57 & 56.65 & 67.30 & 64.59 \\
 &  + Majority Vote & \underline{\textbf{44.93}} & 55.12 & 52.52 & 58.29 & 69.18 & 66.41 \\
  &  + Dis \& Ans & 43.96 & 54.79 & 52.03 & 58.41 & 68.51 & 65.94 \\
  &  + Exp \& Ans & 43.48 & \underline{\textbf{55.78}} & \underline{\textbf{52.64}} & \underline{\textbf{59.57}} & \underline{69.32} & \underline{66.84} \\

\hline
\hline

 \multirow{4}{*}{Phi-3-finetuned}& Context & 44.44 & 51.32 & 49.57 & 56.24 & 64.6 & 62.47 \\
 &  + Majority Vote & 44.44 & 55.94 & 53.01 & 57.26 & 67.94 & 65.22 \\
 & + Dis \& Ans & 38.16 & 49.01 & 46.25 & 49.99 & 61.86 & 58.84 \\
 &  + Exp \& Ans & \underline{\textbf{47.34}} & \underline{\textbf{57.26}} & \underline{\textbf{54.74}} & \underline{\textbf{59.61}} & \underline{\textbf{69.79}} & \underline{\textbf{67.19}} \\
\hline

\end{tabular}
\centering
\caption{Few-shot and finetuned results of models and methods tested on our QACC dataset. EM-C means the Exact Match (EM) score of the set of QA pairs with conflicting contexts, EM-NC means the EM score of QAs with non-conflicting contexts, and EM-T means the total EM score of all the QAs in the test set. Same notation applies to the F1 score. "Context" means LLMs are given both the question and contexts retrieved from google. "+ Majority Vote" means LLMs are given question, contexts and the instruction to take majority vote.
"+ Dis \& Ans" indicates LLMs are given question, contexts, the instruction to discern and answer, and either an in-context example or finetuning data indicating which contexts are perturbed. "+ Exp \& Ans" represents results of LLMs with question, contexts, the instruction to explain and answer, and in-context example of explanation or finetuning data of explanation.
The \textbf{bolded} numbers represent the best results across all few-shot models or finetuned models, and the \underline{underlined} numbers represent the best result in a single model.
}
\label{tab:result1}
\end{table*}

\subsection{Datasets}
We run our experiments on two datasets. The first is the QACC dataset we collect, and the second is a perturbed NQ-Open dataset. For our QACC dataset, we use the validation set to find the best instruction and prompt formats for the LLMs, and report their results on the test set. For the perturbed NQ-Open dataset, we use an entity-substitution method to replace the answer in the contexts to other named entities of the same type in order to create conflicts among the contexts, following \cite{longpre-etal-2021-entity}. We construct this perturbed dataset over the test split of NQ-Open with 3,610 questions. We retrieve top ten results from Google as the contexts for these questions and apply the entity-substitution algorithm with different perturbation ratio. The higher the perturbation ratio means the more contexts in a question are perturbed.

\subsection{Methods}

\paragraph{Retrieval Augmented QA}
Language models can leverage contexts to answer open domain questions \cite{izacard2021leveraging, zhong2022training}. We prompt LLMs with question and contexts retrieved from Google and instruct them that \textit{"Given the following contexts, provide a direct, concise answer to the question"}.

\paragraph{Majority Vote}
As shown in Figure \ref{fig:data_analysis}, humans are inclined to choose the majority answer when there is conflicting evidence. Therefore, we prompt LLMs question and contexts and instruct them to \textit{"use majority vote to decide which context to trust if there are conflicting contexts"}.

\paragraph{Discern and Answer}
\citet{hong2024gullible} proposes to explicitly instruct the model to first discern the counterfactual, perturbed passages and then ignore them to answer the question. We follow the same strategy and instruct the models to \textit{"Find the perturbed passages if there are any, and ignore them when eliciting the correct answer"} with question, contexts, and an example message indicating which of the contexts are "perturbed", using the annotations in QACC that attribute correct/wrong answers to their supporting contexts.

\paragraph{Explain and Answer}
Prompting with explanations introduces richer information that can guide the inference process. Recent work have shown that letting the language model “explain itself” through in-context learning gains more insights into predictions and improves their performances in a variety of reasoning tasks, including question answering \cite{lampinen2022tell, ye2022unreliability, nye2021work, wei2023chainofthought, lampinen2022language}. We believe answering question with conflicting contexts requires similar reasoning abilities and therefore can benefit from eliciting explanations during inference. We instruct the models to \textit{"Explain the reasons and then provide a direct, concise answer"} with question, contexts, and a natural language explanation as in-context example in few-shot and training input in finetuning.
Table \ref{tab:ablation} shows our experiments of comparing Explain then Answer to Answer then Explain. Similar to previous work, we observe only slight impact of the orders of explanation in their performances.

\begin{table*}
\centering
  \begin{tabular}{l|l|SS|SS|SS}
    \toprule
    
    \multirow{3}{*}{Model} & \multirow{3}{*}{Prompt} &
    \multicolumn{6}{c}{Perturbation Ratio} \\
     & & 
      \multicolumn{2}{c}{ 0 \%} &
      \multicolumn{2}{c}{ 25 \%} &
      \multicolumn{2}{c}{ 50 \%} \\
      &  &{EM} & {F1} & {EM} & {F1} & {EM} & {F1} \\
      \midrule

    \multirow{2}{*}{Phi-3}& Zeroshot Context & 15.60 & 33.08 & 14.49 & 31.68 & 13.21 & 29.81 \\
    & Few-shot Context & 30.25 & 44.21 & 28.59 & 42.40 & 25.93 & 39.84 \\
    \hline
    \hline

    \multirow{3}{*}{Phi-3-finetuned}& Context & 34.24 & 47.22 & 32.27 & 45.34 & 29.67 & 42.80 \\
    &  Majority Vote & 34.29 & 46.67& 32.71 & 45.07& 29.72 & 42.31 \\
    &  Dis \& Ans & 32.44 & 44.26 & 30.61 & 42.60& 28.22 & 40.16 \\
    & Exp \& Ans & \textbf{37.31} & \textbf{50.29} & \textbf{35.76} & \textbf{48.57} & \textbf{33.18} & \textbf{46.20} \\
    \bottomrule
  \end{tabular}
  \caption{Zeroshot, few-shot and finetuned results on perturbed NQ-Open test set. The higher the perturbation ratio, the more contexts in a question are perturbed with entity-substitution.}
  \label{tab:nq_perturbed}
\end{table*}

\begin{table}[t]
\begin{tabular} {cc|cc}
\hline
Phi-3 Size & Instruction  & \textsc{EM} & \textsc{F1}  \\

\hline

Mini &\textsc{0-shot-context} & 32.67 & 54.95  \\
Mini &\textsc{1-shot-context} & 47.6 & 63.51  \\
Medium &\textsc{0-shot-context} & 45.14 & 63.45  \\
Medium &\textsc{1-shot-context} & \textbf{49.57} & \textbf{64.59}  \\

\hline
\end{tabular}
\centering
\caption{Zeroshot v.s. few-shot for Phi-3 on QACC. }
\label{tab:zeroshot}
\end{table}


\subsection{Experiment Setup}
 We conduct experiments on three popular instruction-tuned large language models from different families: \textit{GPT-4o-mini}, \textit{Claude3-Sonnet}, and \textit{Phi-3-Medium-Instruct (14B)}, with zero-shot inference, few-shot inference, and finetuning. We find that LLMs greatly benefit from in-context examples (few-shot) compared to zeroshot (See Table \ref{tab:zeroshot}) when answering open domain questions, so we only present few-shot inference and finetuning results in Table \ref{tab:result1}.

For few-shot inference experiments, we include \textbf{one} in-context example of expected input-output pair when prompting the three language models. 
 For finetuning experiments, we finetune Phi-3-Medium-Instruct using LoRA \cite{hu2021lora} with language modeling loss (SFT). We first find the best hyperparameters of finetuning using the validation set of QACC and then train on both the training and validation set and report results in the test set. We also use the validation set of QACC to find the best prompt and instruction format for each methods and use them for both few-shot inference and finetuning. More details of experiment settings are discussed in Appendix \ref{appendix:training}.

 We follow conventions and use Exact Match and F1 scores as the metrics for all our evaluations. The answers in our dataset are predominantly short phrases consisting of only a few words, so we believe that Exact Match and F1 are sufficiently effective in our evaluation. Exact Match returns positive if the generated answer is identical to the reference and negative if otherwise, whereas F1 score is more forgiving and measures the word overlap between the generated and reference answers. We note that LLMs are prone to long generations, so we specifically instruct all of the models to \textit{answer with as few words as possible} in the prompts (see examples of the prompts in Appendix \ref{appendix:prompts}). 

\subsection{Experiment Results}
\paragraph{QACC}
Table \ref{tab:result1} exhibits our experiment results in QACC. We can see that all LLMs that we evaluate inevitably experience worse performance when there are conflicting contexts, comparing their results on EM-C and EM-NC, as well as their results on F1-C and F1-NC. We also find that in different LLMs, their best prompting methods in the few-shot setting are also different. GPT-4o has the best performance when prompted with just the contexts when seeing conflicting contexts, Claude-3 gives the best results when instructed to first explain and then answer the question, and Phi-3 presents comparable performances when instructed to take majority vote and explain then answer. In addition, we perform an additional rule-based context filtering experiment and find that relying solely on context from trusted sources does not yield any improvements (See details in Appendix \ref{appendix:additional_experiments}).

We also demonstrate that by instructing the model to explain its answer and finetuning with our human-written explanations, Phi-3 can improve its performance on question answering with conflicting contexts. We observe an improvement of $2.9\%$ on EM and $3.37\%$ on F1 comparing the models finetuned with just the contexts (Context) and with contexts and the explanations (+Exp \& Ans). Interestingly, we find that by finetuning Phi-3 with contexts and the instruction to take Majority Vote, the model cannot further improve its performance, and finetuning with Discern and Answer instruction and examples hurts the model and diminishes its performance. We hypothesize the reason is that by finetuning with the instruction to take Majority Vote, we are not introducing any new learning signals to the models besides the format, which it already learns from in-context examples, and some QA examples, which Phi-3 may have already seen during its pre-training. On the other hand, finetuning with Discern and Answer data hurts the performance since, although we can attribute the answers to their supporting contexts to create finetuning data for it, our conflicting contexts are naturally existing conflicting information on the web, rather than synthetic perturbed data with only a few entities replaced. This discrepancy re-emphasizes the usefulness of our dataset with naturally conflicting contexts.

\paragraph{Perturbed NQ-Open}

We observe that, similar to real-world scenarios, LLMs tend to perform worse when there are more conflicting contexts in this synthetic dataset. In addition,
the improvements we observe from finetuning Phi-3 on human-written explanations can generalize to perturbed data and general open-domain question answering as well. 
Table \ref{tab:nq_perturbed} exhibits the performance of zeroshot and few-shot Phi-3 models on perturbed NQ-Open, as well as Phi-3 finetuned on QACC and evaluated on perturbed NQ-Open.
As illustrated in Table \ref{tab:nq_perturbed}, Phi-3 finetuned with explanation data consistently outperforms other finetuned models under different ratio of perturbation. We believe that the extra finetuning signals of natural language explanations improve Phi-3's reasoning abilities in general, therefore demonstrating its consistent improvements across all perturbation ratio, including regular open domain QA ($0\%$) and when there are perturbed contexts ($25\%$ and 50$\%$).

\subsection{Analysis}
\begin{figure}
  \includegraphics[width=7.9cm]{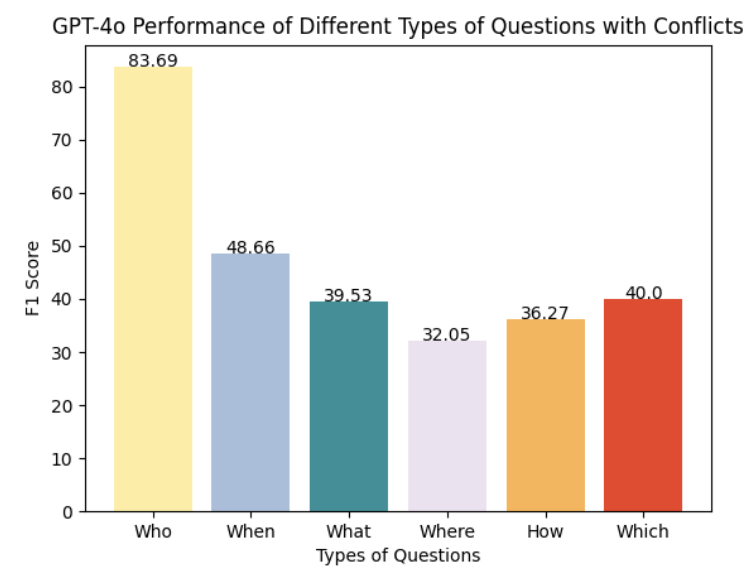}
  \caption{Few-shot GPT-4o performance on the test set of QACC that has conflicting contexts. The x-axis indicates the different types of questions and the y-axis denotes the F1 score for each type.}
  \label{fig:result_analysis}
\end{figure}

Figure \ref{fig:result_analysis} shows the performance of few-shot GPT-4o on the set of questions in QACC that have conflicting contexts. We can see that LLMs like GPT-4o can answer open domain questions reasonably well even with conflicting contexts when the questions are asking about a person (\textit{Who} types of question). However, GPT-4o fails significantly more when the question is asking about a time (\textit{When}), about a place (\textit{Where}), or when the question is open-ended (\textit{How}). We hypothesize that this may relate to the pretraining corpus of LLMs and the frequency that different entities appear in the pretraining corpus: popular people names that appear frequently in LLMs' pretraining corpus allow them to utilize their parametric knowledge to distinguish the answers among the conflicting contexts, whereas the different timestamps and places exist more sparsely in the corpus (as well as on the web), making it more difficult for the LLMs to discern. 

\section{Conclusion}
In this work, we construct a dataset named QACC to study open domain question answering with conflicting contexts. We find that unambiguous, open domain questions are exposed to conflicting evidence on the web: $25\%$ of the questions will lead to conflicting contexts when retrieved using Google, and popular LLMs are very brittle to such conflicts. We show that by finetuning on natural language explanations, we can improve the reasoning abilities of Phi-3 and improve its performances when there are conflicting contexts as well as open domain question answering in general. We will release our dataset and code to promote further research along this line.

\paragraph{Limitations}
Our study and dataset are constrained to factual, unambiguous questions. Specifically, each question in QACC is expected to have a single, definitive factual answer. Subjective questions, unanswerable questions, or questions with multiple answers are beyond the scope of this work. We encourage future research to explore these areas further to gain deeper insights.

For limitations of our data: the pre-defined reasons we ask our annotators to select from may not cover all possible reasons, and these reasons as well as their natural language explanations can be subjective, rather than factual evidence. For evaluation, we adhere to standard conventions by employing Exact Match and F1 scores. While these metrics may be less effective for assessing long, open-ended answers, they are well-suited for our task since the answers in our dataset consist primarily of short phrases.

In addition, eliciting natural language explanations from LLMs have several limitations. Previous work has shown that explanations generated by LLMs can be unreliable and can lead to wrong interpretations of the models. However, in this work, we focus on the improvement of reasoning abilities of LLMs when finetuning with explanation data, rather than interpreting their explanations.

\paragraph{Ethical Statement}
All datasets used in this study are publicly available and comply with relevant licensing and ethical guidelines. No personally identifiable information was processed in our experiments. Our model is designed for research purposes and should not be deployed in high-stakes decision-making without further rigorous evaluation. We discourage any misuse of this technology for generating misleading or harmful content.

\section*{Acknowledgments}
This work was conducted during Siyi Liu’s internship at Amazon Web Services (AWS) and was fully funded by AWS. We sincerely appreciate the valuable feedback provided by the AWS team and extend our gratitude to the ARR reviewers and editors for their insightful and constructive comments.

\bibliography{acl_latex}

\begin{thebibliography}{34}
\providecommand{\natexlab}[1]{#1}

\bibitem[{Borgeaud et~al.(2022)Borgeaud, Mensch, Hoffmann, Cai, Rutherford, Millican, van~den Driessche, Lespiau, Damoc, Clark, de~Las~Casas, Guy, Menick, Ring, Hennigan, Huang, Maggiore, Jones, Cassirer, Brock, Paganini, Irving, Vinyals, Osindero, Simonyan, Rae, Elsen, and Sifre}]{borgeaud2022improving}
Sebastian Borgeaud, Arthur Mensch, Jordan Hoffmann, Trevor Cai, Eliza Rutherford, Katie Millican, George van~den Driessche, Jean-Baptiste Lespiau, Bogdan Damoc, Aidan Clark, Diego de~Las~Casas, Aurelia Guy, Jacob Menick, Roman Ring, Tom Hennigan, Saffron Huang, Loren Maggiore, Chris Jones, Albin Cassirer, Andy Brock, Michela Paganini, Geoffrey Irving, Oriol Vinyals, Simon Osindero, Karen Simonyan, Jack~W. Rae, Erich Elsen, and Laurent Sifre. 2022.
\newblock \href {https://arxiv.org/abs/2112.04426} {Improving language models by retrieving from trillions of tokens}.
\newblock \emph{Preprint}, arXiv:2112.04426.

\bibitem[{Chen et~al.(2017)Chen, Fisch, Weston, and Bordes}]{chen2017reading}
Danqi Chen, Adam Fisch, Jason Weston, and Antoine Bordes. 2017.
\newblock \href {https://arxiv.org/abs/1704.00051} {Reading wikipedia to answer open-domain questions}.
\newblock \emph{Preprint}, arXiv:1704.00051.

\bibitem[{Chen et~al.(2022{\natexlab{a}})Chen, Zhang, and Choi}]{chen-etal-2022-rich}
Hung-Ting Chen, Michael Zhang, and Eunsol Choi. 2022{\natexlab{a}}.
\newblock \href {https://doi.org/10.18653/v1/2022.emnlp-main.146} {Rich knowledge sources bring complex knowledge conflicts: Recalibrating models to reflect conflicting evidence}.
\newblock In \emph{Proceedings of the 2022 Conference on Empirical Methods in Natural Language Processing}, pages 2292--2307, Abu Dhabi, United Arab Emirates. Association for Computational Linguistics.

\bibitem[{Chen et~al.(2019)Chen, Khashabi, Yin, Callison-Burch, and Roth}]{chen2019seeing}
Sihao Chen, Daniel Khashabi, Wenpeng Yin, Chris Callison-Burch, and Dan Roth. 2019.
\newblock Seeing things from a different angle: Discovering diverse perspectives about claims.
\newblock \emph{arXiv preprint arXiv:1906.03538}.

\bibitem[{Chen et~al.(2022{\natexlab{b}})Chen, Liu, Uyttendaele, Zhang, Bruno, and Roth}]{chen2022design}
Sihao Chen, Siyi Liu, Xander Uyttendaele, Yi~Zhang, William Bruno, and Dan Roth. 2022{\natexlab{b}}.
\newblock \href {https://arxiv.org/abs/2112.08357} {Design challenges for a multi-perspective search engine}.
\newblock \emph{Preprint}, arXiv:2112.08357.

\bibitem[{Dasigi et~al.(2021)Dasigi, Lo, Beltagy, Cohan, Smith, and Gardner}]{dasigi2021dataset}
Pradeep Dasigi, Kyle Lo, Iz~Beltagy, Arman Cohan, Noah~A. Smith, and Matt Gardner. 2021.
\newblock \href {https://arxiv.org/abs/2105.03011} {A dataset of information-seeking questions and answers anchored in research papers}.
\newblock \emph{Preprint}, arXiv:2105.03011.

\bibitem[{Guu et~al.(2020)Guu, Lee, Tung, Pasupat, and Chang}]{guu2020realm}
Kelvin Guu, Kenton Lee, Zora Tung, Panupong Pasupat, and Ming-Wei Chang. 2020.
\newblock \href {https://arxiv.org/abs/2002.08909} {Realm: Retrieval-augmented language model pre-training}.
\newblock \emph{Preprint}, arXiv:2002.08909.

\bibitem[{H{\"a}meen-Anttila et~al.(2014)H{\"a}meen-Anttila, Nordeng, Kokki, Jyrkk{\"a}, Lupattelli, Vainio, and Enlund}]{info:doi/10.2196/jmir.2939}
Katri H{\"a}meen-Anttila, Hedvig Nordeng, Esa Kokki, Johanna Jyrkk{\"a}, Angela Lupattelli, Kirsti Vainio, and Hannes Enlund. 2014.
\newblock \href {https://doi.org/10.2196/jmir.2939} {Multiple information sources and consequences of conflicting information about medicine use during pregnancy: A multinational internet-based survey}.
\newblock \emph{J Med Internet Res}, 16(2):e60.

\bibitem[{Han et~al.(2021)Han, Hsu, Sun, Baylon, Ning, Roth, and Peng}]{han-etal-2021-ester}
Rujun Han, I-Hung Hsu, Jiao Sun, Julia Baylon, Qiang Ning, Dan Roth, and Nanyun Peng. 2021.
\newblock \href {https://doi.org/10.18653/v1/2021.emnlp-main.597} {{ESTER}: A machine reading comprehension dataset for reasoning about event semantic relations}.
\newblock In \emph{Proceedings of the 2021 Conference on Empirical Methods in Natural Language Processing}, pages 7543--7559, Online and Punta Cana, Dominican Republic. Association for Computational Linguistics.

\bibitem[{Hong et~al.(2024)Hong, Kim, Kang, Myaeng, and Whang}]{hong2024gullible}
Giwon Hong, Jeonghwan Kim, Junmo Kang, Sung-Hyon Myaeng, and Joyce~Jiyoung Whang. 2024.
\newblock \href {https://arxiv.org/abs/2305.01579} {Why so gullible? enhancing the robustness of retrieval-augmented models against counterfactual noise}.
\newblock \emph{Preprint}, arXiv:2305.01579.

\bibitem[{Hu et~al.(2021)Hu, Shen, Wallis, Allen-Zhu, Li, Wang, Wang, and Chen}]{hu2021lora}
Edward~J. Hu, Yelong Shen, Phillip Wallis, Zeyuan Allen-Zhu, Yuanzhi Li, Shean Wang, Lu~Wang, and Weizhu Chen. 2021.
\newblock \href {https://arxiv.org/abs/2106.09685} {Lora: Low-rank adaptation of large language models}.
\newblock \emph{Preprint}, arXiv:2106.09685.

\bibitem[{Izacard and Grave(2021)}]{izacard2021leveraging}
Gautier Izacard and Edouard Grave. 2021.
\newblock \href {https://arxiv.org/abs/2007.01282} {Leveraging passage retrieval with generative models for open domain question answering}.
\newblock \emph{Preprint}, arXiv:2007.01282.

\bibitem[{Karpukhin et~al.(2020)Karpukhin, Oğuz, Min, Lewis, Wu, Edunov, Chen, and tau Yih}]{karpukhin2020dense}
Vladimir Karpukhin, Barlas Oğuz, Sewon Min, Patrick Lewis, Ledell Wu, Sergey Edunov, Danqi Chen, and Wen tau Yih. 2020.
\newblock \href {https://arxiv.org/abs/2004.04906} {Dense passage retrieval for open-domain question answering}.
\newblock \emph{Preprint}, arXiv:2004.04906.

\bibitem[{Kasai et~al.(2023)Kasai, Sakaguchi, takahashi, Le~Bras, Asai, Yu, Radev, Smith, Choi, and Inui}]{NEURIPS2023_9941624e}
Jungo Kasai, Keisuke Sakaguchi, yoichi takahashi, Ronan Le~Bras, Akari Asai, Xinyan Yu, Dragomir Radev, Noah~A Smith, Yejin Choi, and Kentaro Inui. 2023.
\newblock \href {https://proceedings.neurips.cc/paper_files/paper/2023/file/9941624ef7f867a502732b5154d30cb7-Paper-Datasets_and_Benchmarks.pdf} {Realtime qa: What\textquotesingle s the answer right now?}
\newblock In \emph{Advances in Neural Information Processing Systems}, volume~36, pages 49025--49043. Curran Associates, Inc.

\bibitem[{Kazemi et~al.(2023)Kazemi, Yuan, Bhatia, Kim, Xu, Imbrasaite, and Ramachandran}]{kazemi2023boardgameqa}
Mehran Kazemi, Quan Yuan, Deepti Bhatia, Najoung Kim, Xin Xu, Vaiva Imbrasaite, and Deepak Ramachandran. 2023.
\newblock \href {https://arxiv.org/abs/2306.07934} {Boardgameqa: A dataset for natural language reasoning with contradictory information}.
\newblock \emph{Preprint}, arXiv:2306.07934.

\bibitem[{Khandelwal et~al.(2020)Khandelwal, Levy, Jurafsky, Zettlemoyer, and Lewis}]{khandelwal2020generalization}
Urvashi Khandelwal, Omer Levy, Dan Jurafsky, Luke Zettlemoyer, and Mike Lewis. 2020.
\newblock \href {https://arxiv.org/abs/1911.00172} {Generalization through memorization: Nearest neighbor language models}.
\newblock \emph{Preprint}, arXiv:1911.00172.

\bibitem[{Lampinen et~al.(2022{\natexlab{a}})Lampinen, Dasgupta, Chan, Matthewson, Tessler, Creswell, McClelland, Wang, and Hill}]{lampinen2022language}
Andrew~K. Lampinen, Ishita Dasgupta, Stephanie C.~Y. Chan, Kory Matthewson, Michael~Henry Tessler, Antonia Creswell, James~L. McClelland, Jane~X. Wang, and Felix Hill. 2022{\natexlab{a}}.
\newblock \href {https://arxiv.org/abs/2204.02329} {Can language models learn from explanations in context?}
\newblock \emph{Preprint}, arXiv:2204.02329.

\bibitem[{Lampinen et~al.(2022{\natexlab{b}})Lampinen, Roy, Dasgupta, Chan, Tam, McClelland, Yan, Santoro, Rabinowitz, Wang, and Hill}]{lampinen2022tell}
Andrew~K. Lampinen, Nicholas~A. Roy, Ishita Dasgupta, Stephanie C.~Y. Chan, Allison~C. Tam, James~L. McClelland, Chen Yan, Adam Santoro, Neil~C. Rabinowitz, Jane~X. Wang, and Felix Hill. 2022{\natexlab{b}}.
\newblock \href {https://arxiv.org/abs/2112.03753} {Tell me why! explanations support learning relational and causal structure}.
\newblock \emph{Preprint}, arXiv:2112.03753.

\bibitem[{Lee et~al.(2019)Lee, Chang, and Toutanova}]{lee-etal-2019-latent}
Kenton Lee, Ming-Wei Chang, and Kristina Toutanova. 2019.
\newblock \href {https://doi.org/10.18653/v1/P19-1612} {Latent retrieval for weakly supervised open domain question answering}.
\newblock In \emph{Proceedings of the 57th Annual Meeting of the Association for Computational Linguistics}, pages 6086--6096, Florence, Italy. Association for Computational Linguistics.

\bibitem[{Liu et~al.(2023)Liu, Lin, Hewitt, Paranjape, Bevilacqua, Petroni, and Liang}]{liu2023lostmiddlelanguagemodels}
Nelson~F. Liu, Kevin Lin, John Hewitt, Ashwin Paranjape, Michele Bevilacqua, Fabio Petroni, and Percy Liang. 2023.
\newblock \href {https://arxiv.org/abs/2307.03172} {Lost in the middle: How language models use long contexts}.
\newblock \emph{Preprint}, arXiv:2307.03172.

\bibitem[{Liu et~al.(2021)Liu, Chen, Uyttendaele, and Roth}]{liu2021multioped}
Siyi Liu, Sihao Chen, Xander Uyttendaele, and Dan Roth. 2021.
\newblock Multioped: A corpus of multi-perspective news editorials.
\newblock \emph{arXiv preprint arXiv:2106.02725}.

\bibitem[{Liu et~al.(2024)Liu, Yao, Lv, Fan, Cao, Yu, Hou, and Li}]{liu2024untangle}
Yantao Liu, Zijun Yao, Xin Lv, Yuchen Fan, Shulin Cao, Jifan Yu, Lei Hou, and Juanzi Li. 2024.
\newblock \href {https://arxiv.org/abs/2404.03577} {Untangle the knot: Interweaving conflicting knowledge and reasoning skills in large language models}.
\newblock \emph{Preprint}, arXiv:2404.03577.

\bibitem[{Longpre et~al.(2021)Longpre, Perisetla, Chen, Ramesh, DuBois, and Singh}]{longpre-etal-2021-entity}
Shayne Longpre, Kartik Perisetla, Anthony Chen, Nikhil Ramesh, Chris DuBois, and Sameer Singh. 2021.
\newblock \href {https://doi.org/10.18653/v1/2021.emnlp-main.565} {Entity-based knowledge conflicts in question answering}.
\newblock In \emph{Proceedings of the 2021 Conference on Empirical Methods in Natural Language Processing}, pages 7052--7063, Online and Punta Cana, Dominican Republic. Association for Computational Linguistics.

\bibitem[{Min et~al.(2020)Min, Michael, Hajishirzi, and Zettlemoyer}]{min2020ambigqa}
Sewon Min, Julian Michael, Hannaneh Hajishirzi, and Luke Zettlemoyer. 2020.
\newblock \href {https://arxiv.org/abs/2004.10645} {Ambigqa: Answering ambiguous open-domain questions}.
\newblock \emph{Preprint}, arXiv:2004.10645.

\bibitem[{Ning et~al.(2020)Ning, Wu, Dasigi, Dua, Gardner, Logan~IV, Marasovi{\'c}, and Nie}]{ning-etal-2020-easy}
Qiang Ning, Hao Wu, Pradeep Dasigi, Dheeru Dua, Matt Gardner, Robert~L. Logan~IV, Ana Marasovi{\'c}, and Zhen Nie. 2020.
\newblock \href {https://doi.org/10.18653/v1/2020.emnlp-demos.17} {Easy, reproducible and quality-controlled data collection with {CROWDAQ}}.
\newblock In \emph{Proceedings of the 2020 Conference on Empirical Methods in Natural Language Processing: System Demonstrations}, pages 127--134, Online. Association for Computational Linguistics.

\bibitem[{Nye et~al.(2021)Nye, Andreassen, Gur-Ari, Michalewski, Austin, Bieber, Dohan, Lewkowycz, Bosma, Luan, Sutton, and Odena}]{nye2021work}
Maxwell Nye, Anders~Johan Andreassen, Guy Gur-Ari, Henryk Michalewski, Jacob Austin, David Bieber, David Dohan, Aitor Lewkowycz, Maarten Bosma, David Luan, Charles Sutton, and Augustus Odena. 2021.
\newblock \href {https://arxiv.org/abs/2112.00114} {Show your work: Scratchpads for intermediate computation with language models}.
\newblock \emph{Preprint}, arXiv:2112.00114.

\bibitem[{Pan et~al.(2023)Pan, Chen, Kan, and Wang}]{pan2023attacking}
Liangming Pan, Wenhu Chen, Min-Yen Kan, and William~Yang Wang. 2023.
\newblock \href {https://arxiv.org/abs/2110.07803} {Attacking open-domain question answering by injecting misinformation}.
\newblock \emph{Preprint}, arXiv:2110.07803.

\bibitem[{Voorhees and Tice(2000)}]{voorhees-tice-2000-trec}
Ellen~M. Voorhees and Dawn~M. Tice. 2000.
\newblock \href {http://www.lrec-conf.org/proceedings/lrec2000/pdf/26.pdf} {The {TREC}-8 question answering track}.
\newblock In \emph{Proceedings of the Second International Conference on Language Resources and Evaluation ({LREC}{'}00)}, Athens, Greece. European Language Resources Association (ELRA).

\bibitem[{Wan et~al.(2024)Wan, Wallace, and Klein}]{wan2024evidence}
Alexander Wan, Eric Wallace, and Dan Klein. 2024.
\newblock \href {https://arxiv.org/abs/2402.11782} {What evidence do language models find convincing?}
\newblock \emph{Preprint}, arXiv:2402.11782.

\bibitem[{Wei et~al.(2023)Wei, Wang, Schuurmans, Bosma, Ichter, Xia, Chi, Le, and Zhou}]{wei2023chainofthought}
Jason Wei, Xuezhi Wang, Dale Schuurmans, Maarten Bosma, Brian Ichter, Fei Xia, Ed~Chi, Quoc Le, and Denny Zhou. 2023.
\newblock \href {https://arxiv.org/abs/2201.11903} {Chain-of-thought prompting elicits reasoning in large language models}.
\newblock \emph{Preprint}, arXiv:2201.11903.

\bibitem[{Xie et~al.(2024)Xie, Zhang, Chen, Lou, and Su}]{xie2024adaptive}
Jian Xie, Kai Zhang, Jiangjie Chen, Renze Lou, and Yu~Su. 2024.
\newblock \href {https://arxiv.org/abs/2305.13300} {Adaptive chameleon or stubborn sloth: Revealing the behavior of large language models in knowledge conflicts}.
\newblock \emph{Preprint}, arXiv:2305.13300.

\bibitem[{Ye and Durrett(2022)}]{ye2022unreliability}
Xi~Ye and Greg Durrett. 2022.
\newblock \href {https://arxiv.org/abs/2205.03401} {The unreliability of explanations in few-shot prompting for textual reasoning}.
\newblock \emph{Preprint}, arXiv:2205.03401.

\bibitem[{Zhang and Choi(2021)}]{SituatedQA}
Michael J.~Q. Zhang and Eunsol Choi. 2021.
\newblock \href {https://arxiv.org/abs/2109.06157} {Situatedqa: Incorporating extra-linguistic contexts into qa}.
\newblock \emph{Preprint}, arXiv:2109.06157.

\bibitem[{Zhong et~al.(2022)Zhong, Lei, and Chen}]{zhong2022training}
Zexuan Zhong, Tao Lei, and Danqi Chen. 2022.
\newblock \href {https://arxiv.org/abs/2205.12674} {Training language models with memory augmentation}.
\newblock \emph{Preprint}, arXiv:2205.12674.

\end{thebibliography}

\appendix




\section{Experiment Details}
\label{appendix:training}
We perform our finetuning experiments using Amazon Elastic Compute Cloud (Amazon EC2). We use one p4d.24xlarge instance for the training. It has 8 NVIDIA A100 GPUs with 40.0 GB GPU memory each. We use Parameter-Efficient Fine-Tuning (PEFT) and Low-Rank Adaptation (LoRA) to train our models. We use learning rata 2e-4, lora\_r = 16, lora\_alpha = 32, and lora\_dropout = 0.05.

We use snippets returned by Google Search as the contexts for each of the articles retrieved. Studies have shown that LLMs struggle to process and understand very long contexts \cite{liu2023lostmiddlelanguagemodels}, so we believe that Google snippets are good summarization and extraction of their original articles for our model to process different contexts. To guarantee that the answer always exists in the contexts, we remove examples that our annotators believe they didn't find a correct answer in the contexts in our experiments. Examples of our annotation interface are shown in Appendix \ref{appendix:annotation}.

\section{Additional Experiment Results}
\label{appendix:additional_experiments}
As suggested by one of our ARR reviewers, we perform a rule-based context filtering experiment to understand whether exclusively using contexts from trusted sources can be a simple solution. Specifically, we only include contexts from our list of trusted sources, i.e., Wikipedia, URLs with top-level government domain extensions, and URLs with .edu top-level domain extensions as the contexts for each question. There are about $5\%$ of questions do not have any context from these sources, and we include all original contexts for these questions. As shown in Table \ref{tab:rulebased}, we observe that a simple context filtering baseline does not yield any improvements. One potential reason is that it is difficult to define a thorough list of trusted sources that can apply to all open domain questions spanning different domains like entertainment, history, and science, etc. This finding highlights again the challenge of building conflict-aware systems that can distinguish across conflicting evidence.

\begin{table}[]
\begin{tabular} {c|ccc}
\hline
Prompt  & \textsc{F1-C} & \textsc{F1-NC} & \textsc{F1-T}  \\

\hline
Context & \textbf{59.26} & \textbf{70.06} & \textbf{67.32}  \\
Trusted Context  & 57.47 & 63.57 & 62.02 \\

\hline
\end{tabular}
\centering
\caption{GPT4o-mini few-shot experiments on rule-based, context filtering baseline.}
\label{tab:rulebased}
\end{table}

\begin{figure*}[h]
  \includegraphics[width=\textwidth]{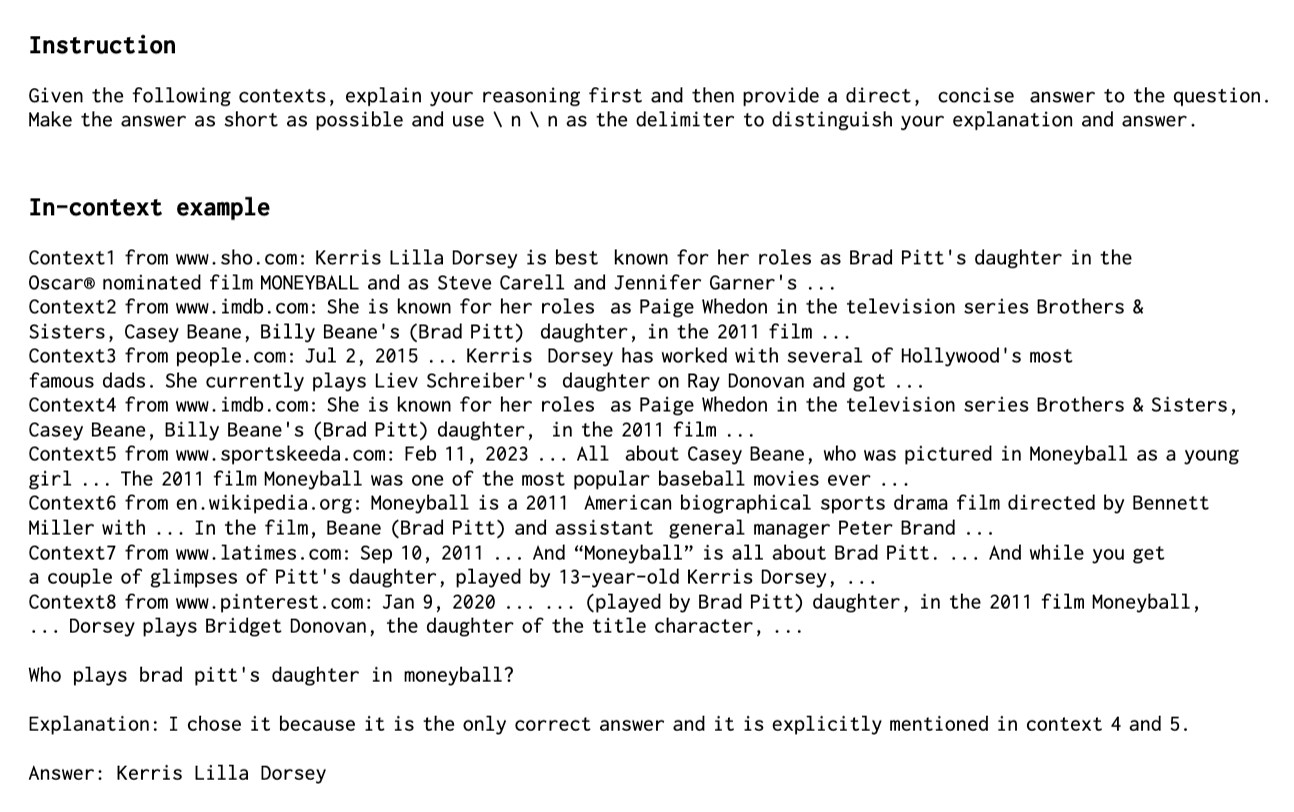}
  \caption{One example of our prompts}
  \label{fig:prompts}
\end{figure*}

\section{Example Prompt}
\label{appendix:prompts}
We present an example of the prompt and in-context example we use for running our experiments in Figure \ref{fig:prompts}. We preserve the same format and instruction in our finetuning experiments for consistency.

\section{Human Annotation Examples}
\label{appendix:annotation}

\begin{figure*}
  \includegraphics[width=\textwidth]{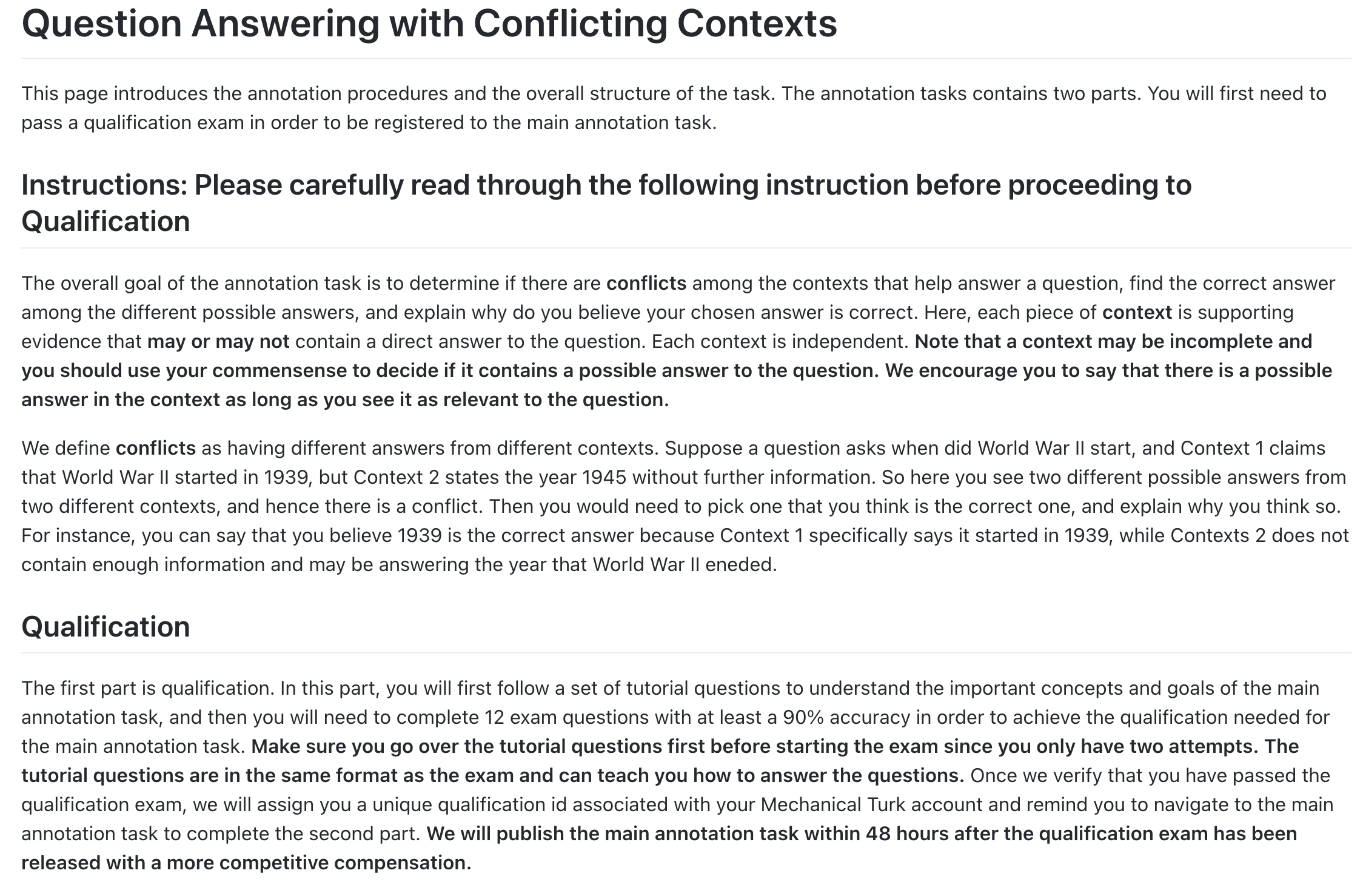}
  \caption{Instruction example 1.}
  \label{fig:instruction1}
\end{figure*}

\begin{figure*}
  \includegraphics[width=\textwidth]{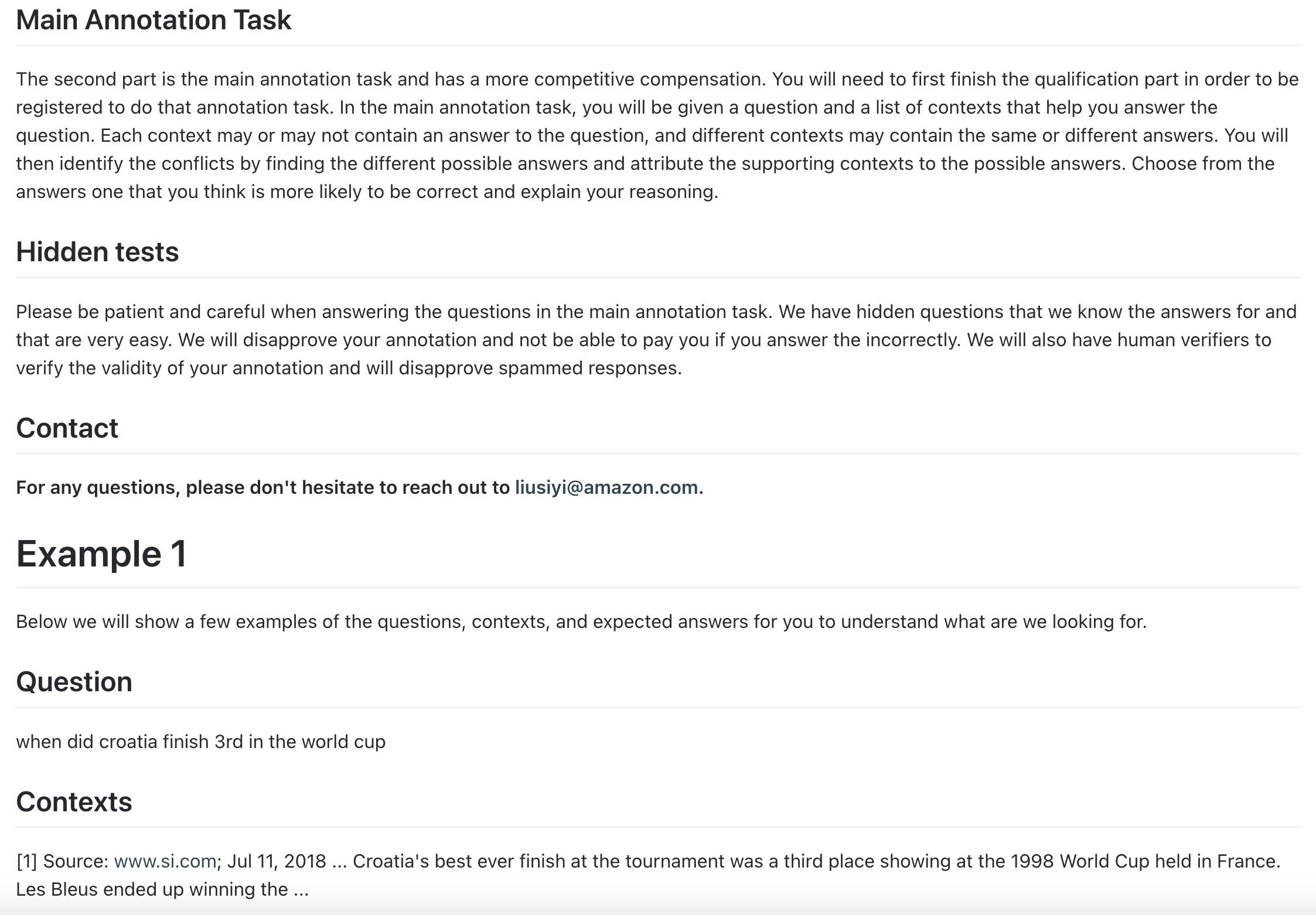}
  \caption{Instruction example 2.}
  \label{fig:instruction2}
\end{figure*}

\begin{figure*}
  \includegraphics[width=\textwidth]{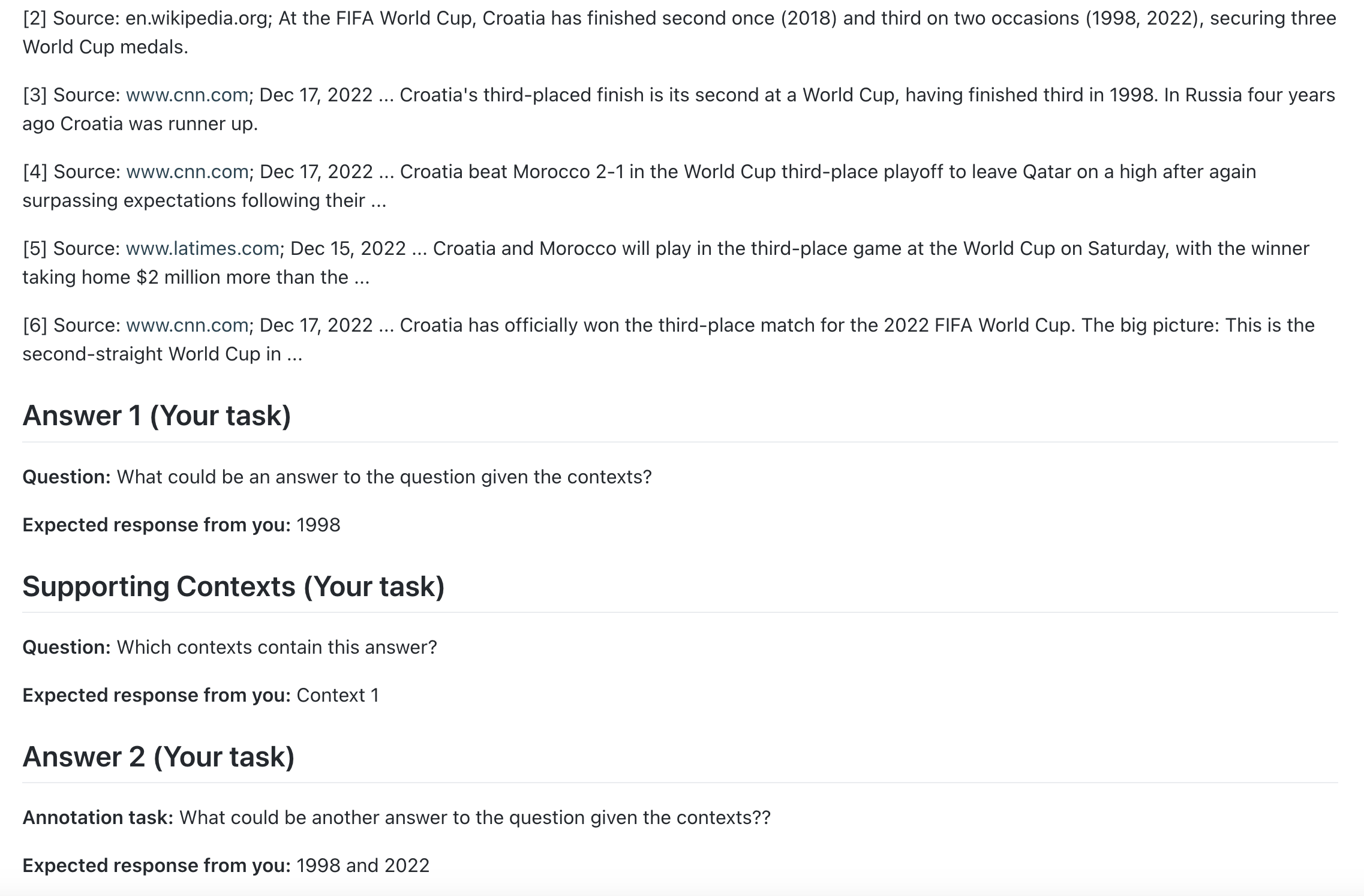}
  \caption{Instruction example 3.}
  \label{fig:instruction3}
\end{figure*}

\begin{figure*}
  \includegraphics[width=\textwidth]{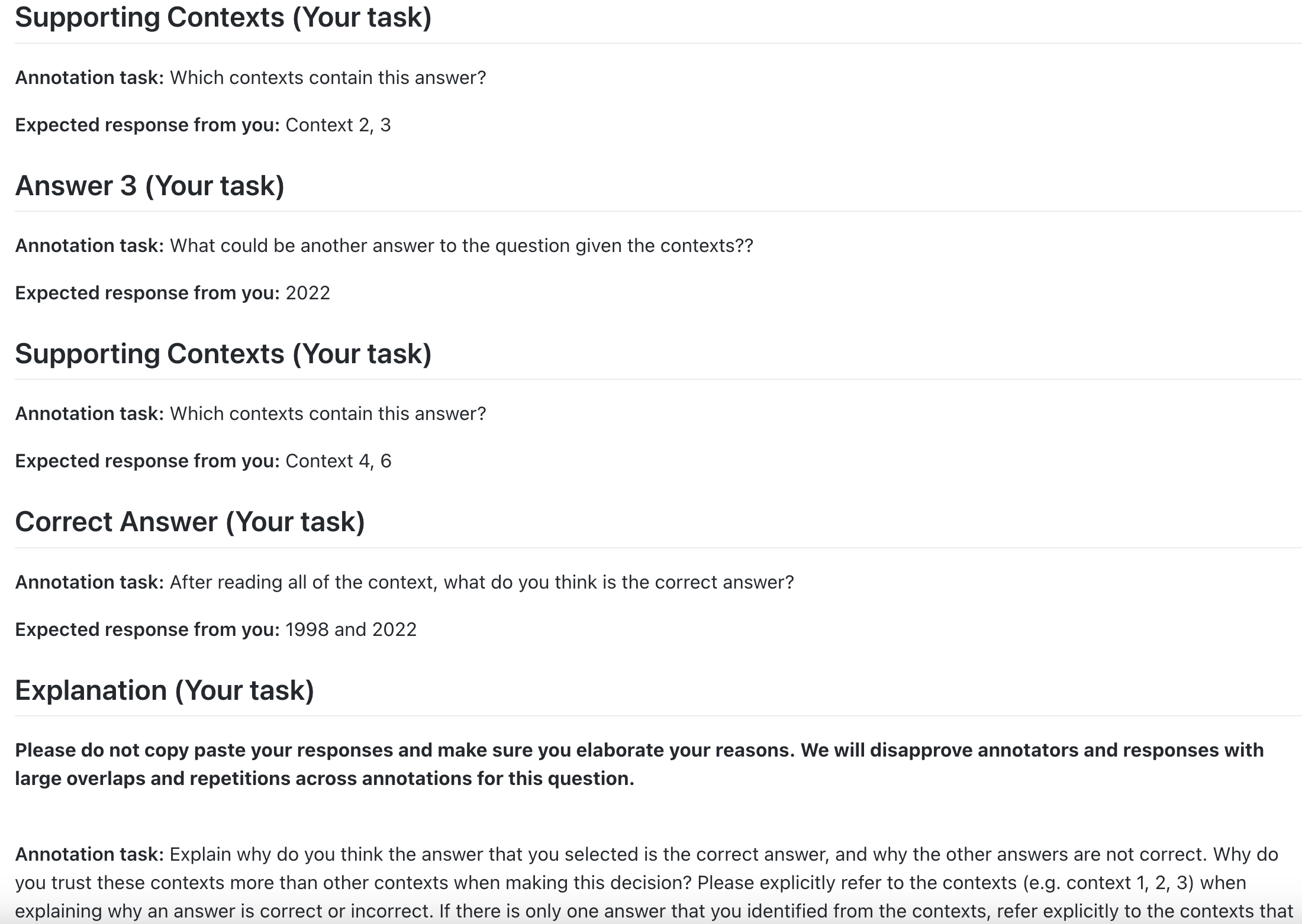}
  \caption{Instruction example 4.}
  \label{fig:instruction4}
\end{figure*}

\begin{figure*}
  \includegraphics[width=\textwidth]{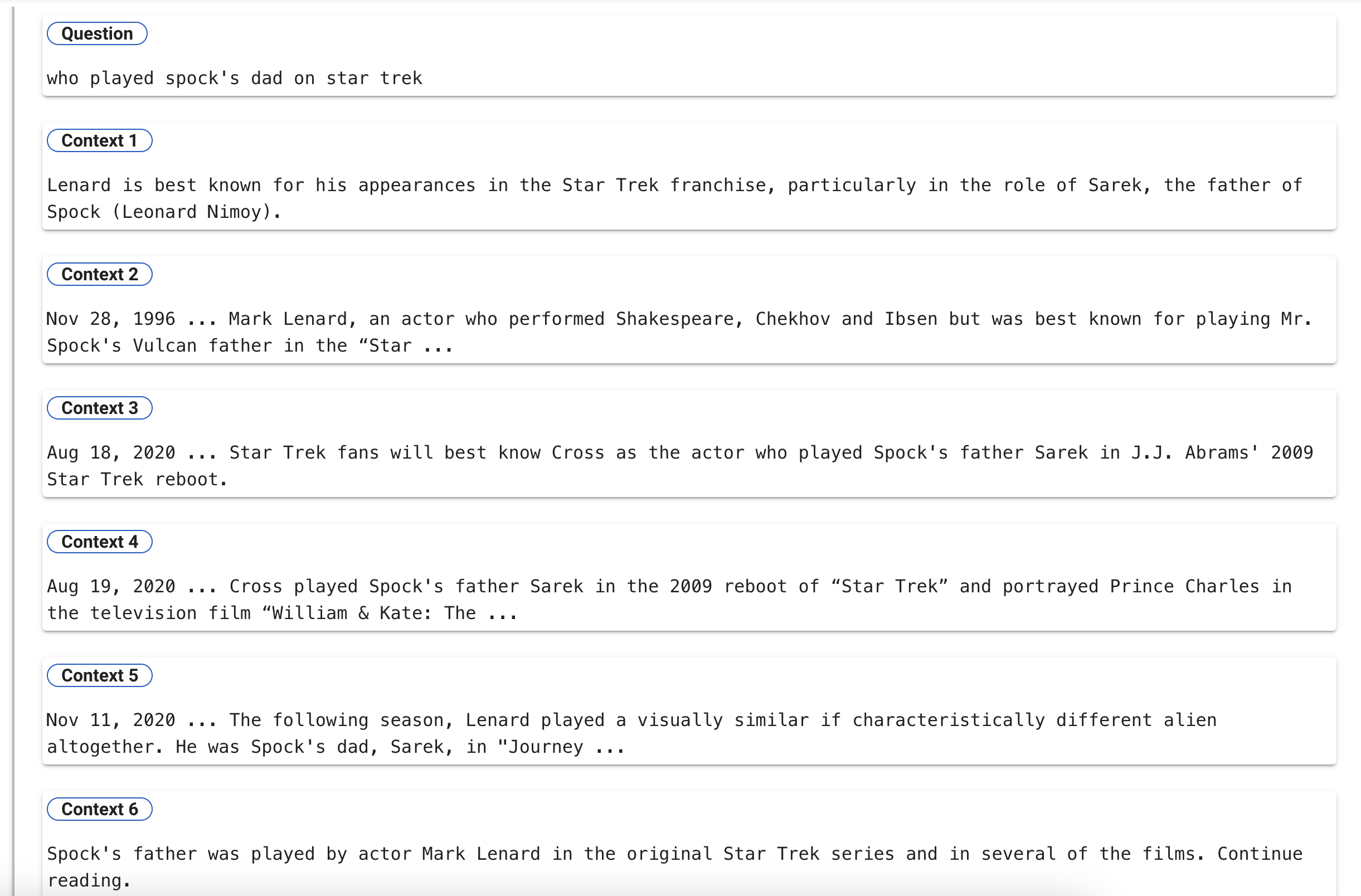}
  \caption{Qualification example 1.}
  \label{fig:qualification1}
\end{figure*}

\begin{figure*}
  \includegraphics[width=\textwidth]{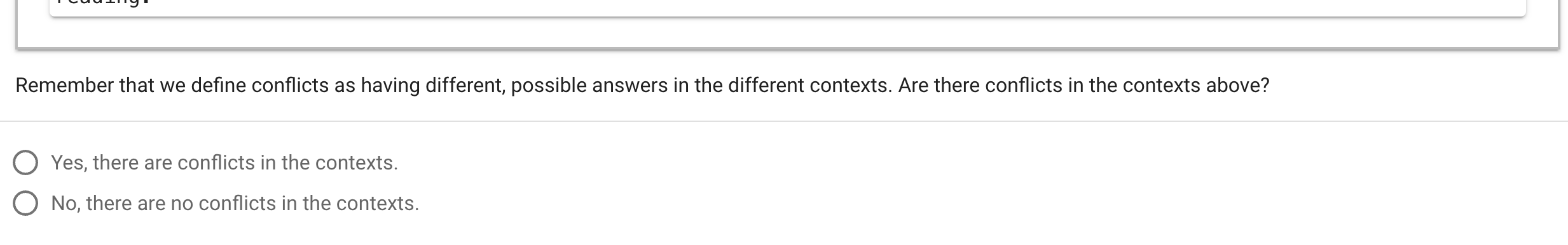}
  \caption{Qualification example 2.}
  \label{fig:qualification2}
\end{figure*}

\begin{figure*}
  \includegraphics[width=\textwidth]{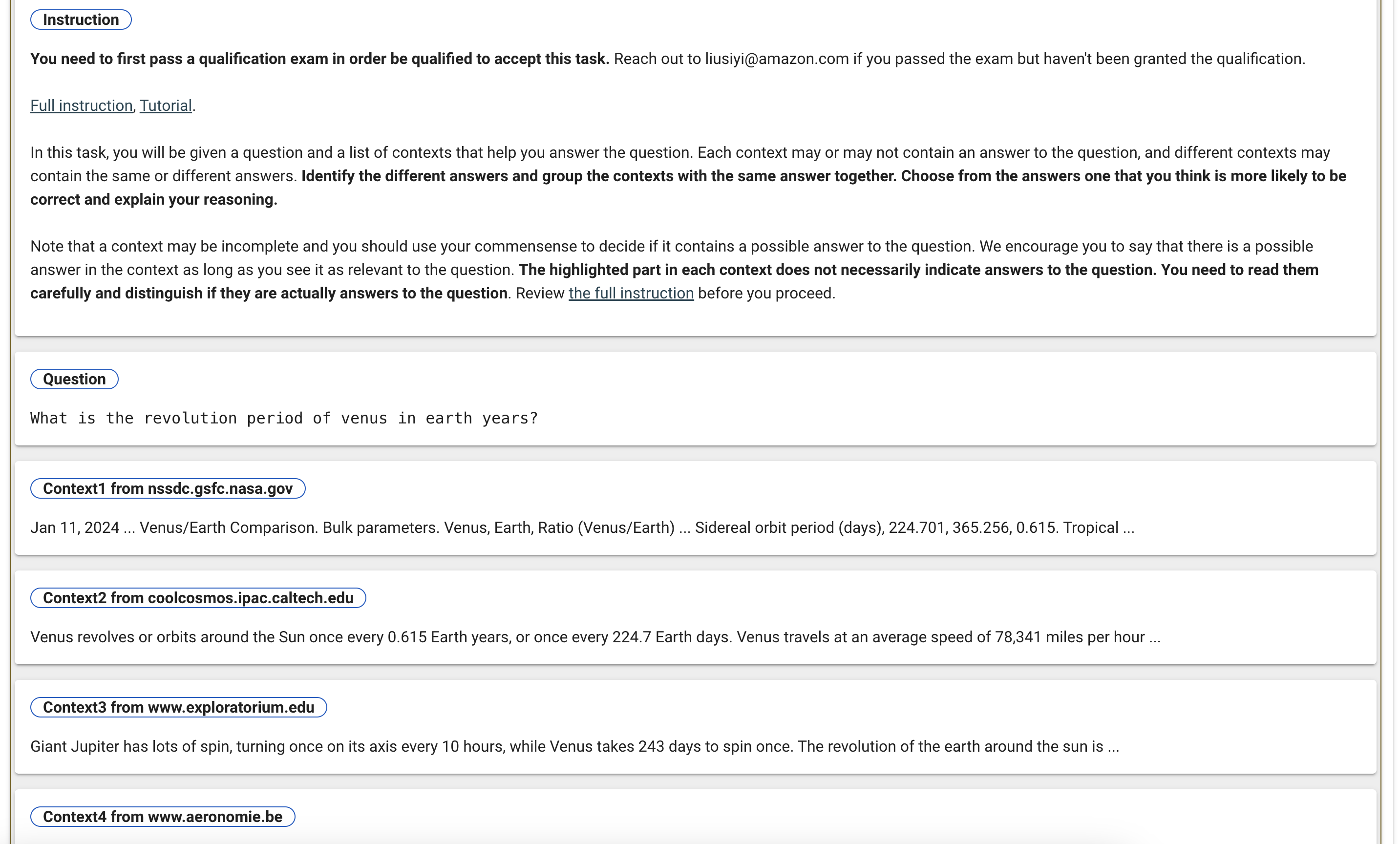}
  \caption{Annotation example 1.}
  \label{fig:annotation1}
\end{figure*}

\begin{figure*}
  \includegraphics[width=\textwidth]{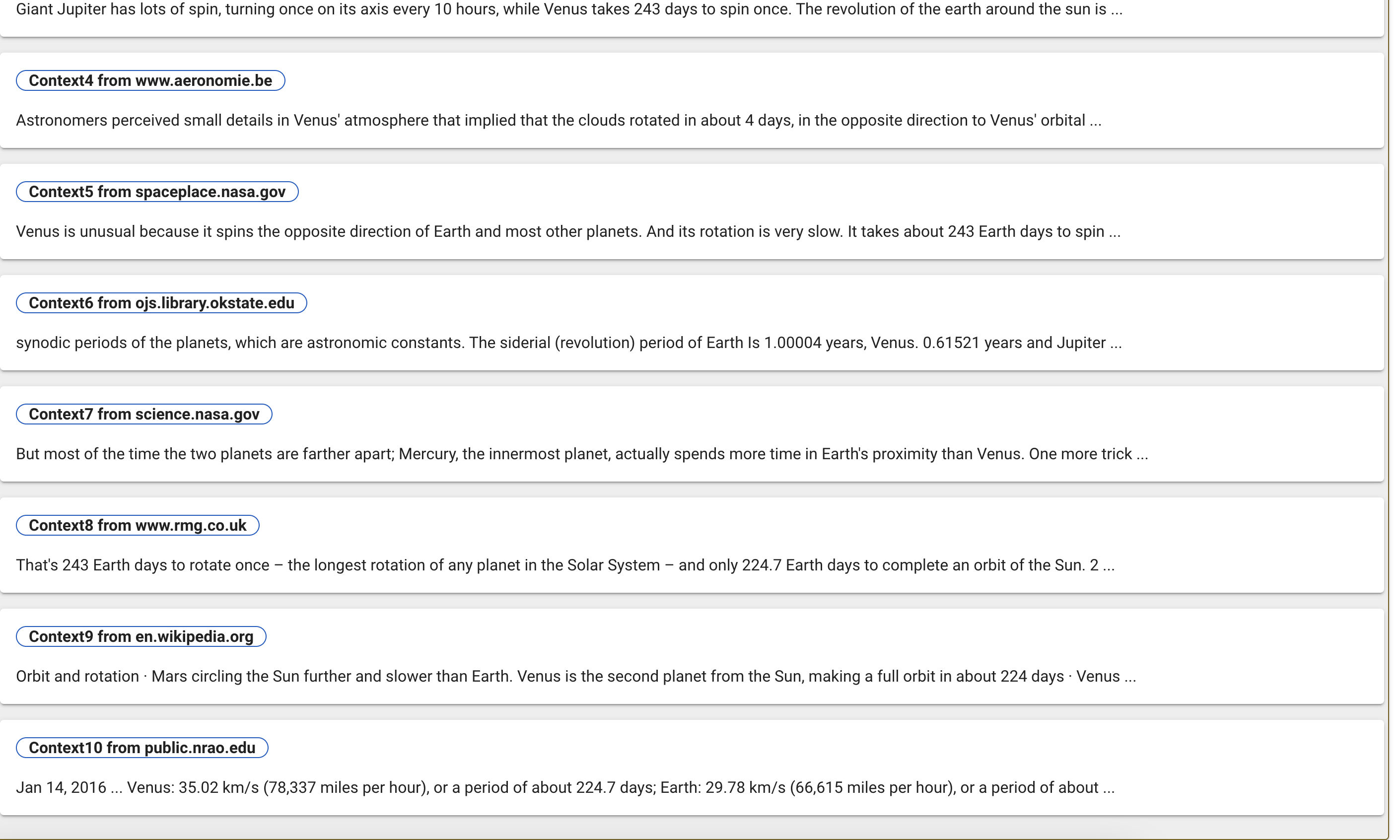}
  \caption{Annotation example 2.}
  \label{fig:annotation2}
\end{figure*}

\begin{figure*}
  \includegraphics[width=\textwidth]{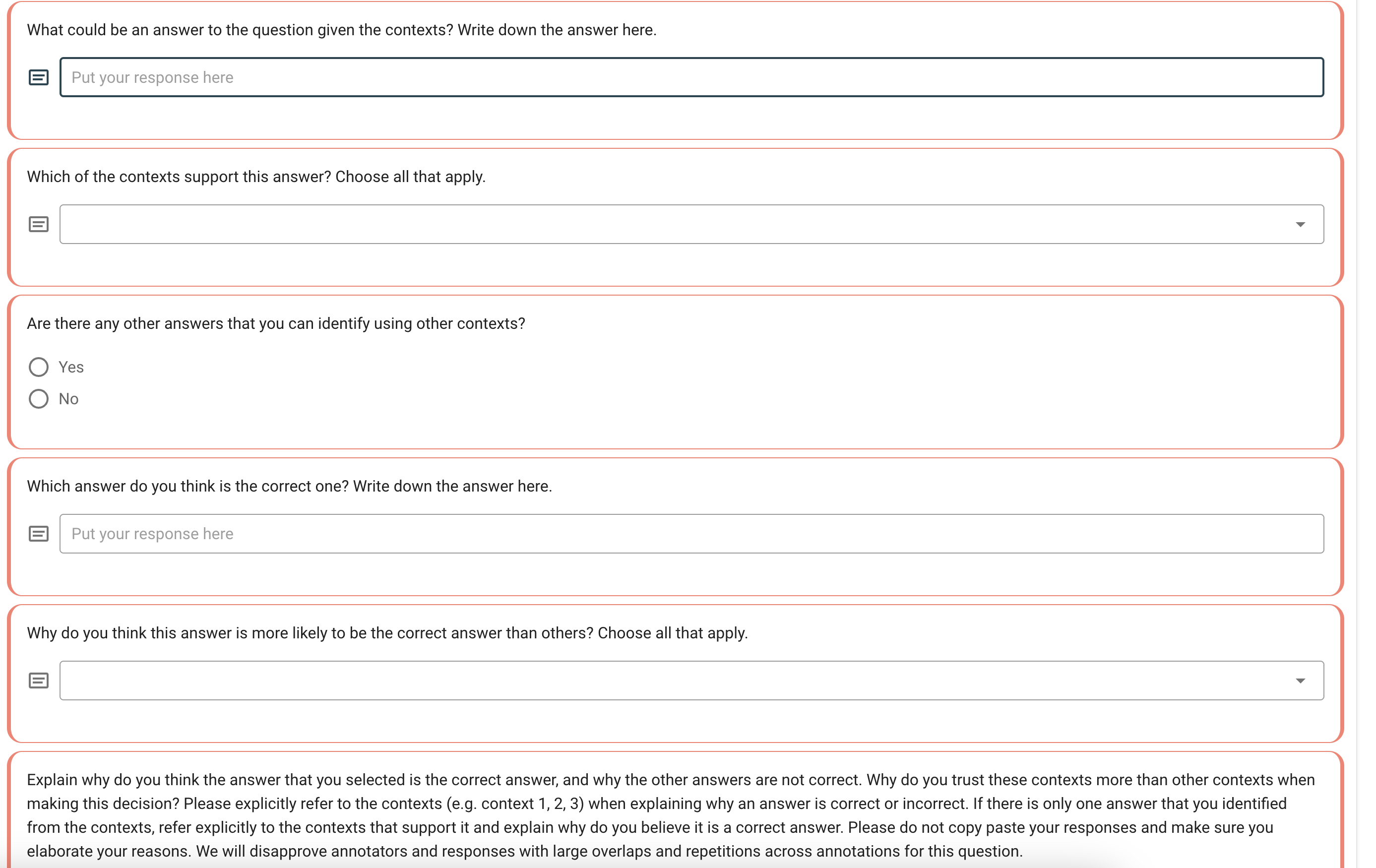}
  \caption{Annotation example 3.}
  \label{fig:annotation3}
\end{figure*}

\end{document}